\pdfoutput=1

\documentclass[11pt]{article}

\usepackage[]{acl}

\usepackage{times}
\usepackage{latexsym}

\usepackage{graphicx}
\usepackage{amsfonts}
\usepackage{amsmath}

\usepackage{booktabs}

\usepackage[T1]{fontenc}

\usepackage[utf8]{inputenc}

\usepackage{microtype}

\title{Fusing Heterogeneous Factors with Triaffine Mechanism \\ for Nested Named Entity Recognition}

\author{
Zheng Yuan$^{12}$\thanks{$\quad$Work done at Alibaba DAMO Academy.} \space\space\space
Chuanqi Tan$^{2}$ \space\space
Songfang Huang$^{2}$ \space\space
Fei Huang$^{2}$\\
$^{1}$Tsinghua University \space\space\space\space
$^{2}$Alibaba Group\\
\texttt{yuanz17@mails.tsinghua.edu.cn}\\
\texttt{\{chuanqi.tcq,songfang.hsf,f.huang\}@alibaba-inc.com}
}

\begin{document}
\maketitle
\begin{abstract}
Nested entities are observed in many domains due to their compositionality, which cannot be easily recognized by the widely-used sequence labeling framework.
A natural solution is to treat the task as a span classification problem.
To learn better span representation and increase classification performance, it is crucial to effectively integrate heterogeneous factors including inside tokens, boundaries, labels, and related spans which could be contributing to nested entities recognition.
To fuse these heterogeneous factors, we propose a novel triaffine mechanism including triaffine attention and scoring.
Triaffine attention uses boundaries and labels as queries and uses inside tokens and related spans as keys and values for span representations.
Triaffine scoring interacts with boundaries and span representations for classification.
Experiments show that our proposed method outperforms previous span-based methods, achieves the state-of-the-art $F_1$ scores on nested NER datasets GENIA and KBP2017, and shows comparable results on ACE2004 and ACE2005.
\end{abstract}

\section{Introduction}

Named entity recognition (NER) is a fundamental natural language processing task that extracts entities from texts.
Flat NER has been well studied and is usually viewed as a sequence labeling problem \cite{lample2016neural}. However, nested entities also widely exist in real-world applications due to their multi-granularity semantic meaning \cite{alex2007recognising, yuan2020unsupervised},
which cannot be solved by the sequence labeling framework since tokens have multiple labels \cite{finkel2009nested}.

\begin{figure}[t]
\centering
\includegraphics[width=2.4in]{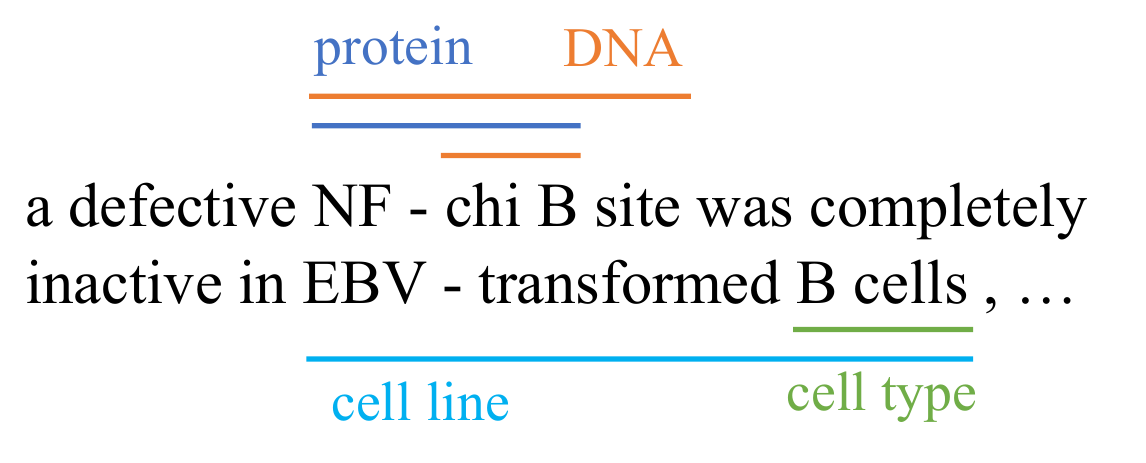}
\caption{An example sentence with nested entities from the GENIA dataset.}
\label{fig:example}
\end{figure}

Various paradigms for nested NER have been proposed in recent years. A representative direction is the span-based approach that learns deep representation for every possible span and then classifies it to the corresponding type \cite{zheng2019boundary,mgner,wadden2019entity,bensc,wang2020pyramid,yu2020named}. 
By leveraging the large-scale pretrained language model, several works show that the simple model structure for span representation and classification can achieve satisfactory results \cite{dygie,zhong2021frustratingly}. However, we still believe that explicit modeling of some relevant features will further benefit the span representation and classification under the complex nested setting.
Taking Figure \ref{fig:example} as an example, we claim that the following factors are critical for recognizing whether a span is an entity.
(1) \textbf{Tokens}: 
It is obvious that tokens of the given span contribute to the recognition.
(2) \textbf{Boundaries}: 
We emphasize boundaries (or boundary tokens) because they are special tokens with rich semantics. Works with simple structure may just produce the span representation based on the concatenation or biaffine transformation of boundary representation \cite{yu2020named,fu2021nested}. Some other works take boundary detection as additional supervision for better representation learning \cite{zheng2019boundary,bensc}. More importantly, a unilateral boundary cannot determine the entity type since it can exist in multiple entities with different labels (e.g., ``\textit{NF}'', ``\textit{B}'', and ``\textit{cells}'') under the nested setting.
(3) \textbf{Labels}:
As mentioned above, tokens could belong to entities with different labels.
Therefore, we propose that the model should learn label-aware span representation to take into consideration of the different token contributions at the label level.\footnote{Label is the perdition object that we cannot touch in representation learning. Here, leveraging label information only means we need label-aware representation learning.}
For example, ``\textit{NF}'' may contribute more to ``protein'' type when classifying the span ``\textit{NF - chi B}'', as well as ``\textit{chi B}'' and ``\textit{site}'' contribute more to ``DNA'' type when classifying the span ``\textit{NF - chi B site}''.
(4) \textbf{Related spans}:
Interactions among spans are important in nested entities \cite{luo2020bipartite, wang2020pyramid, fu2021nested}.
The insider and outsider entities may hint at each other's types.
For example, entities inside ``\textit{EBV-transformed B cells}'' have more possibilities to be cell-related entities.
Interactions can also help the non-entity span like ``\textit{transformed B cells}'' to validate its partialness by looking at outer entity ``\textit{EBV - transformed B cells}''.

Although some of the factors may have been explored in previous works, to the best of our knowledge, it is the first work to fuse all these heterogeneous factors into a unified network. As the traditional additive, multiplicative attention, or biaffine transformation cannot interact with such multiple heterogeneous factors simultaneously, we propose a novel triaffine mechanism as the tensor multiplication with three rank-1 tensors (vectors) and a rank-3 tensor, which makes it possible to jointly consider high-order interactions among multiple factors. Specifically, our method follows the pipeline of span representation learning and classification. At the stage of span representation learning, we apply the triaffine attention to aggregate the label-wise span representations by considering boundaries and labels as queries as well as inside tokens as keys and values. Then, a similar triaffine attention is applied to produce the label-wise cross-span representations by querying boundaries and labels with related spans. At the stage of span classification, we fuse the span representations and boundaries for label-wise classification with a triaffine score function. 
In practice, we add an auxiliary object function to classify spans without the cross-span interaction, which benefits learning robust span representation and can be used as a span filter to speed up both training and inference without performance degradation.

We conduct experiments on four nested NER datasets: ACE2004, ACE2005, GENIA, and KBP2017. 
Our model achieves 88.56, 88.83, 81.23, and 87.27 scores in terms of $F_1$, respectively.
Using the BERT encoder, our model outperforms state-of-the-art methods on GENIA and KBP2017 and shows comparable performances on ACE2004 and ACE2005 with the latest generative methods.
Ablation studies show the effectiveness of each factor and the superiority of the triaffine mechanism. 

Our contributions are summarized as \footnote{Codes and models are available at \url{https://github.com/GanjinZero/Triaffine-nested-ner}.}:
\begin{itemize}
    \item We propose that heterogeneous factors (i.e., tokens, boundaries, labels, related spans) should be taken into consideration in the span-based methods for nested NER.
    \item We propose a span-based method with a novel triaffine mechanism including triaffine attention and scoring to fuse the above-mentioned heterogeneous factors for span representations and classification.
    \item Experiments show that our proposed method performs better than existing span-based methods and achieves state-of-the-arts performances on GENIA and KBP17.
\end{itemize}

\section{Related Work}
\subsection{Nested NER}
Nested NER approaches do not have a unified paradigm.
Here we mainly focus on span-based methods since they are close to our work.

The span-based methods are one of the most mainstream ways for the nested NER. With the development of pre-training, it is easy to obtain the span representation by the concatenation of boundary representation \cite{dygie,zhong2021frustratingly} or the aggregated representation of tokens \cite{zheng2019boundary,wadden2019entity}, and then follow a linear layer \cite{mgner} or biaffine transformation \cite{yu2020named} for classification. Several works improve the span-based methods with additional features or supervision. \citet{zheng2019boundary,bensc} point out the importance of boundaries and therefore introduce the boundary detection task.
\citet{wang2020pyramid} propose Pyramid to allow interactions between spans from different layers.
\citet{fu2021nested} adopt TreeCRF to model interactions between nested spans.
Compared with previous methods, our method can jointly fuse multiple heterogeneous factors with the proposed triaffine mechanism.

Other methods for nested NER vary greatly. Earlier research on nested NER is rule-based \cite{zhang2004enhancing}. \citet{lu2015joint,katiyar2018nested,sh} leverage the hypergraph to represent all possible nested structures, which needs to be carefully designed to avoid spurious structures and structural ambiguities.
\citet{trans,mergelabel} predict the transition actions to construct nested entities. \citet{lin2019sequence} propose an anchor-based method to recognize entities. There are other works that recognize entities in a generative fashion \cite{bartner,shen2021locate,tan2021sequence}. Generally, it is not a unified framework for nested NER, and we model it with a span-based method since it is most straightforward.

\begin{figure*}[ht]
\centering
\includegraphics[width=4.5in]{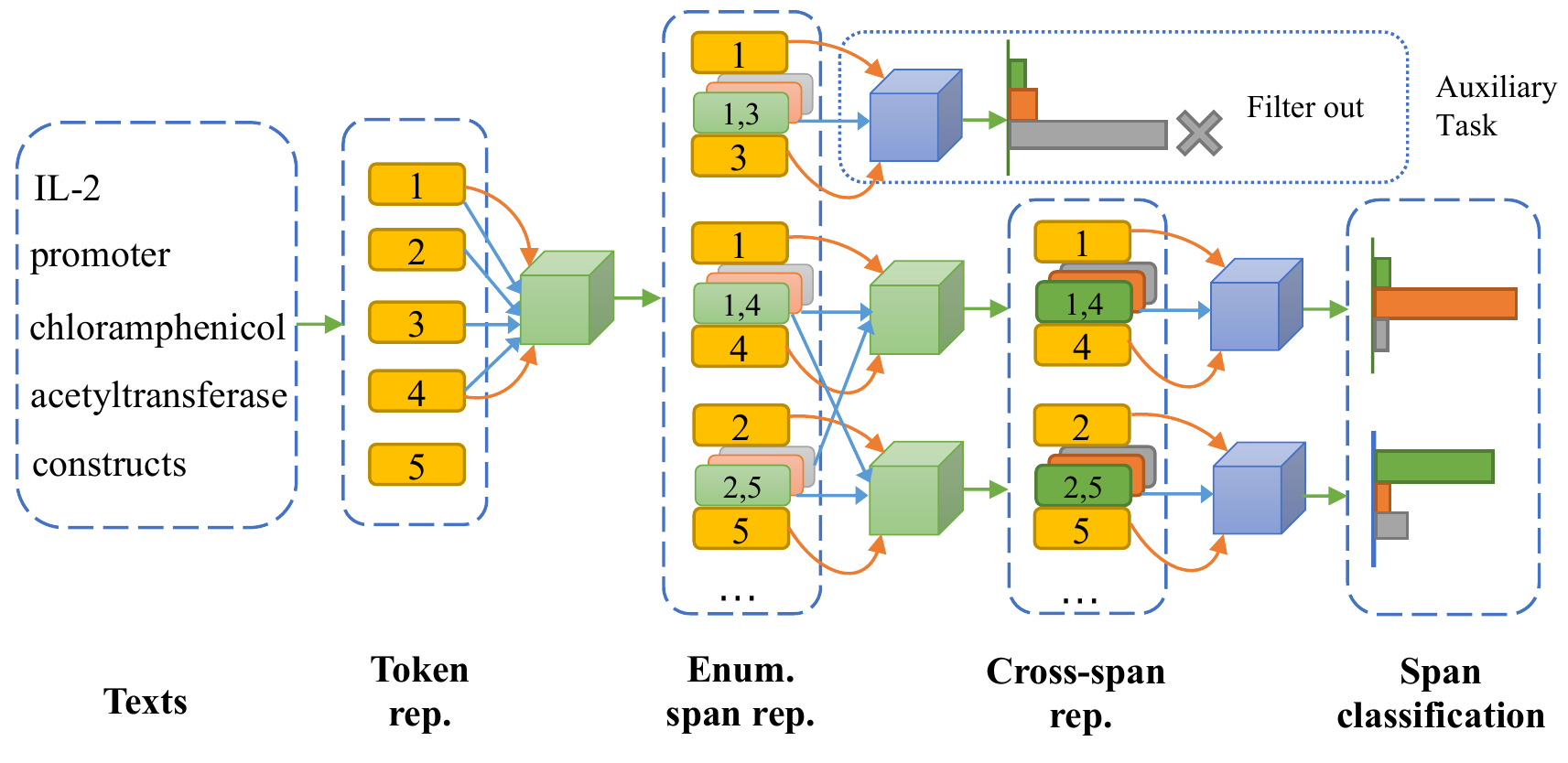}
\caption{The architecture of our method.
Green cubes indicate triaffine attention. Blue cubes indicate triaffine scoring.
Orange arrows mean boundary information. 
Blue arrows mean inside tokens or related spans information.
For each span, we have head and tail representations in yellow and label-wise span representations in different colors.
The grey color indicates None class.
}
\label{arch}
\centering
\includegraphics[width=5.8in]{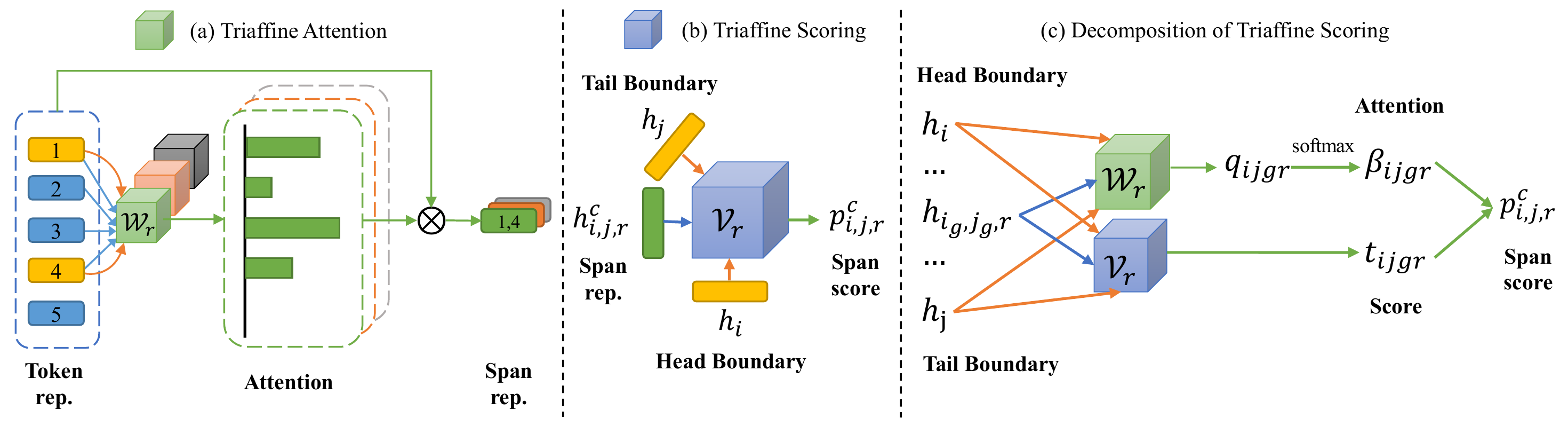}
\caption{Visualization of triaffine attention, triaffine scoring, and the decomposition of triaffine scoring.}
\label{tri}
\end{figure*}

\subsection{Affine Transformations in NLP}
\citet{dozat2016deep} introduce the biaffine transformation in the dependency parsing task for arc classification. Later, it is widely used in many tasks that need to model bilateral representations \cite{li2019dependency,yu2020named}.
The triaffine transformation is further introduced to extend biaffine transformation for high-order interaction in the field of dependency parsing \cite{wang2019second,zhang2020efficient} and semantic role labeling \cite{li2020high}. Except for the similar formula of vectors' interactions, the motivation and the use of triaffine are different in our paper.
Firstly, they only model the homogeneous features such as three tokens, but our triaffine transformation can model heterogeneous factors including labels, boundaries, and related spans. Secondly, they usually leverage triaffine transformation to obtain log potentials for CRFs, but we apply it for span representation and classification.


\section{Approach}

Figure~\ref{arch} shows an overview of our method. We will first introduce the triaffine transformations, which lie in the heart of our model to fuse heterogeneous factors. Then, we will introduce our model including triaffine attention and triaffine scoring based on the proposed triaffine transformations.


\subsection{Deep Triaffine Transformation}
We define the deep triaffine transformation with vectors $\mathbf{u}, \mathbf{v}, \mathbf{w} \in \mathbb{R}^{d}$ and a tensor $\mathcal W \in \mathbb{R}^{d+1}\times\mathbb{R}^{d}\times\mathbb{R}^{d+1}$ which outputs a scalar by applying distinct MLP (multi-layer perceptron) transformations on input vectors and calculating tensor vector multiplications. A constant $1$ is concatenated with inputs to retain the biaffine transformation.
\begin{align}
    \mathbf{u}' = \left[
    	\begin{aligned}
    	    \mathbf{{\rm MLP_a}(u)} \\
    	    1
    	\end{aligned}
    	\right]&,
    \mathbf{v}' = \left[
    	\begin{aligned}
    	    \mathbf{{\rm MLP_c}(v)} \\
    	    1
    	\end{aligned}
    	\right] \\
    \mathbf{w}'=&\mathbf{{\rm MLP_b}(w)}  \\
    {\rm TriAff}(\mathbf{u}, \mathbf{v}, \mathbf{w}, {\mathcal W})=&
    	{\mathcal W} \times_{1} \mathbf{u}' \times_{2} \mathbf{w'} \times_{3} \mathbf{v}'
    \label{eq:tri}
\end{align}
where $\times_{n}$ is the mode-$n$ tensor vector multiplication and ${\rm MLP_t}$ is a $t$-layer MLP (0-layer MLP is equal to identify function).
The tensor $\mathcal W$ is initialized using $\mathcal N(0,\sigma^2)$.
In our approach, we use boundary representations as $\mathbf{u}$ and $\mathbf{v}$. Inside tokens or span representations are used as $\mathbf{w}$.
We denote the tensors in the triaffine attention as $\{\mathcal W_r\}$ and triaffine scoring as $\{\mathcal V_r\}$, which decouples attention weights and scores for different labels.

\subsection{Text Encoding}

We follow \citet{ju-etal-2018-neural,shen2021locate,tan2021sequence} to encode the text. For text $X = [x_1, x_2, ..., x_N]$ with $N$ tokens, we first generate the contextual embedding $\mathbf{x}_{i}^{c}$ with the pre-trained language model,
\begin{align}
    \mathbf{x}_{1}^{c}, \mathbf{x}_{2}^{c}, ..., \mathbf{x}_{N}^{c} = {\rm PLM}(x_1, x_2, ..., x_N)
\end{align}
Then, we concatenate $\mathbf{x}_{i}^{c}$ with word embedding $\mathbf{x}_i^{w}$, part-of-speech embedding $\mathbf{x}_i^{p}$ and character embedding $\mathbf{x}_i^{ch}$, and feed the concatenated embedding $\mathbf{x}_{i}$ into a BiLSTM \cite{hochreiter1997long} to obtain the token representations $\{\mathbf{h}_{i}\}$.

\subsection{Triaffine Attention for Span Representations} \label{triaffine}

To fuse heterogeneous factors for better span representation, we propose a triaffine attention mechanism shown in Figure~\ref{tri}a. 
To interact tokens with labels and boundaries, we learn the label-wise span representation $\mathbf{h}_{i,j,r}$ with the triaffine attention $\alpha_{i,j,k,r}$ for the span $(i,j)$:
\begin{align}
    s_{i,j,k,r} & = {\rm TriAff}(\mathbf{h}_i, \mathbf{h}_j, \mathbf{h}_k, {\mathcal W}_r) \\
    \alpha_{i,j,k,r} & = \frac{\exp(s_{i,j,k,r})}{\sum_{k'=i}^j \exp (s_{i,j,k',r})} \\
    \mathbf{h}_{i,j,r} & = \sum_{k=i}^j\alpha_{i,j,k,r}{\rm MLP}(\mathbf{h}_k)
\label{attn0}
\end{align}
Boundary representations ($\mathbf{h}_i$, $\mathbf{h}_j$) and the label-wise parameters (${\mathcal W}_r$) can be viewed as attention queries, and tokens ($\mathbf{h}_k$) can be viewed as keys and values.
Compared with the general attention framework (additive or multiplicative attention), our triaffine attention permits all high-order interactions between heterogeneous queries and keys.




\subsection{Triaffine Attention for Cross-span Representations}
Motivated by the span-level interactions in the nested setting,
we fuse related spans information 
into cross-span representations.
We view the boundaries of the span and labels as attention queries, related spans (containing the span itself) as attention keys and values to obtain cross-span representations.
Similar to the Equation~\ref{attn0}, we obtain label-wise cross-span representations $\mathbf{h}^c_{i,j,r}$ for the span $(i,j)$ based on triaffine attention $\beta_{i,j,g,r}$.
\begin{align}
    q_{i,j,g,r} &= {\rm TriAff}(\mathbf{h}_{i}, \mathbf{h}_{j}, \mathbf{h}_{i_g,j_g,r}, {\mathcal W}_r) \\
    \beta_{i,j,g,r} &= \frac{\exp(q_{i,j,g,r})}{\sum_{g'}\exp(q_{i,j,g',r})} \\
    \mathbf{h}^c_{i,j,r} &= \sum_g\beta_{i,j,g,r}{\rm MLP}(\mathbf{h}_{i_g,j_g,r})
\end{align}
where $\{(i_g,j_g)\}$ are the related spans. One can treat all enumerated spans as related spans, and we will introduce how we select them in Section \ref{s3.6}.

\subsection{Triaffine Scoring for Span Classification}
To classify the entity type of the span, we calculate label-wise scores based on cross-span representations.
Since boundary information has been proved effective in previous works \cite{yu2020named,fu2021nested}, 
we leverage the boundaries information and cross-span representations for span classification via triaffine scoring. 
Specifically, we estimate the log probabilities $p_{i,j,r}^c$ of the span $(i,j)$ for label $r$ using boundaries $\mathbf{h}_{i},\mathbf{h}_{j}$ and cross-span representations $\mathbf{h}^c_{i,j,r}$.
\begin{equation}
    p_{i,j,r}^c = {\rm TriAff}(\mathbf{h}_i, \mathbf{h}_j, \mathbf{h}^c_{i,j,r}, {\mathcal V}_r)
    \label{eq:slow0}
\end{equation}
Since $\mathbf{h}^c_{i,j,r}$ are composed by $\mathbf{h}_{i_g,j_g,r}$, we can decompose Equation~\ref{eq:slow0} into following if and only if the layer of MLP transformation on $\mathbf{h}^c_{i,j,r}$ is 0:
\begin{align}
    t_{i,j,g,r} &= {\rm TriAff}(\mathbf{h}_{i}, \mathbf{h}_{j}, \mathbf{h}_{i_g,j_g,r}, {\mathcal V}_r) \\
    p^c_{i,j,r} &= \sum_g\beta_{i,j,g,r}t_{i,j,g,r}
\end{align}
Figure~\ref{tri}b and~\ref{tri}c show the mechanism of triaffine scoring and the decomposition. We also apply the similar decomposition functions in the auxiliary span classification task, which applies the triaffine scoring on boundary representations and intermediate span representations $\mathbf{h}_{i,j,r}$ to estimate log probabilities $p_{i,j,r}$ as intermediate predictions.


\subsection{Training and Inference}
\label{s3.6}
In practice, it is expensive and non-informative to consider interactions between all spans. Therefore, we propose an auxiliary task to classify spans with intermediate span representations.
Then, we can rank all spans based on the maximum of log probabilities (except None) from the intermediate predictions $p_{i,j}=\max_{r=1}^Rp_{i,j,r}$, and retain top-$m$ spans $\{(i_l,j_l)\}_{l=1}^m$ as candidates.
We calculate cross-span representations $\mathbf{h}_{i_l,j_l,r}^c$ for retained spans by considering the full interactions among them, and estimate the classification logits $p_{i_l,j_l,r}^c$.
Thus, we have two groups of predictions in our model $\{p_{i,j,r}\}_{1\leq i\leq j \leq N}$ and $\{p^c_{i_l,j_l,r}\}_{1\leq l \leq m}$.
$\{p_{i,j,r}\}$ are calculated for every possible span, and $\{p^{c}_{i_l,j_l,r}\}$ are calculated only on top-$m$ spans.

In the training phase, we jointly minimize two groups of cross-entropy losses:
\begin{align}
     \mathcal L_{aux} =& -\frac{2}{N(N+1)}\sum_{i,j} \log \frac{\exp(p_{i,j,r_{ij}})}{\sum_r\exp(p_{i,j,r})} \\
     \mathcal L_{main} =& -\frac{1}{m}\sum_{1 \leq l \leq m} \log \frac{\exp(p^c_{i_l,j_l,r_{i_l,j_l}})}{\sum_r\exp(p^c_{i_l,j_l,r})} \\
    \mathcal L =& \mu_{aux}\mathcal L_{aux} + \mathcal L_{main}
\end{align}
where $r_{ij}$ is the label of span $(i,j)$.

In both the training and inference phase, $\{p_{i,j,r}\}$ are used to select spans with high possibilities based on the supervision from $\mathcal L_{aux}$.
We inference the labels of selected spans using $\{p^{c}_{i_l,j_l,r}\}$ by assigning label $\tilde{r}_{i_l,j_l}=\arg_r\max p^{c}_{i_l,j_l,r}$, and we assign None class for others.

\section{Experiments}

\begin{table*}[t]
    \small 
    \centering
    \begin{tabular}{lccccccccccc}
    \toprule
       & \multicolumn{3}{c}{ACE2004} & \multicolumn{3}{c}{ACE2005} & \multicolumn{2}{c}{GENIA} & \multicolumn{3}{c}{KBP2017} \\
       & Train & Dev & Test & Train & Dev & Test & Train & Test & Train & Dev & Test \\
    \midrule
    $\#$ sentences & 6,200 & 745 & 812 & 7,194 & 969 & 1,047 & 16,692 & 1,854 & 10,546 & 545 & 4,267\\
    $\#$ entities & 22,204 & 2,514 & 3,035 & 24,411 & 3,200 & 2,993 & 50,509 & 5,506 & 31,236 & 1,879 & 12,601 \\
    $\#$ nested entities & 10,149 & 1,092 & 1,417 & 9,389 & 1,112 & 1,118 & 9,064 & 1,199 & 8,773 & 605 & 3,707 \\
    max entity count & 28 & 22 & 20 & 27 & 23 & 17 & 25 & 14 & 58 & 15 & 21 \\
    \bottomrule
    \end{tabular}
    \caption{Statistics of nested NER datasets ACE2004, ACE2005, GENIA, and KBP2017.}
    \label{dataset}
\end{table*}

\begin{table*}[ht]
\small
\centering
\begin{tabular}{lccccccccc}
\toprule
Model + Encoder & \multicolumn{3}{c}{ACE2004} & \multicolumn{3}{c}{ACE2005} & \multicolumn{3}{c}{GENIA} \\
& P & R & $F_1$ & P & R & $F_1$& P & R & $F_1$  \\
\midrule
\bf{Span-based Methods} \\
DYGIE \cite{dygie} + LSTM & - & - & 84.7 & - & - & 82.9 & - & - & 76.2 \\
MGNER \cite{mgner} + ELMo & 81.7 & 77.4 & 79.5 & 79.0 & 77.3 & 78.2 & - & - & -  \\
BENSC \cite{bensc} & 85.8 & 84.8 & 85.3 & 83.8 & 83.9 & 83.9 & 79.2 & 77.4 & 78.3  \\
TreeCRF \cite{fu2021nested}  & 86.7 & 86.5 & 86.6 & 84.5 & 86.4 & 85.4 & 78.2 & 78.2 & 78.2 \\
Biaffine \cite{yu2020named}  & 87.3 & 86.0 & 86.7 & 85.2 & 85.6 & 85.4 & 81.8 & 79.3 & 80.5 \\
Pyramid \cite{wang2020pyramid}  & 86.08 & 86.48 & 86.28 & 83.95 & 85.39 & 84.66 & 79.45 & 78.94 & 79.19 \\
Pyramid \cite{wang2020pyramid} + ALBERT & 87.71 & 87.78 & 87.74 & 85.30 & 87.40 & 86.34 & 80.33 & 78.31 & 79.31 \\
\midrule
\bf{Others} \\
SH \cite{sh} + LSTM & 78.0 & 72.4 & 75.1 & 76.8 & 72.3 & 74.5 & 77.0 & 73.3 & 75.1 \\
ARN \cite{lin2019sequence} + LSTM & 76.2 & 73.6 & 74.9 & 75.8 & 73.9 & 74.8 & - & - & - \\
BiFlag \cite{luo2020bipartite} + LSTM & - & - & - & 75.0 & 75.2 & 75.1 & 77.4 & 74.6 & 76.0 \\
Merge Label \cite{mergelabel} & - & - & - & 82.7 & 82.1 & 82.4 & - & - & - \\
Seq2seq \cite{strakova2019neural} & - & - & 84.40 & - & - & 84.33 & - & - & 78.31 \\
Second-best \cite{second} & 85.94 & 85.69 & 85.82 & 83.83 & 84.87 & 84.34 & 77.81 & 76.94 & 77.36 \\
BartNER \cite{bartner} + BART  & 87.27 & 86.41 & 86.84 & 83.16 & 86.38 & 84.74 & 78.87 & 79.60 & 79.23 \\
Sequence to Set \cite{tan2021sequence}   & 88.46 & 86.10 & 87.26 & 87.48 & 86.63 & 87.05 & 82.31 & 78.66 & 80.44 \\
Locate and Label \cite{shen2021locate}  & 87.44 & 87.38 & 87.41 & 86.09 & 87.27 & 86.67 & 80.19 & 80.89 & 80.54 \\
\midrule
{\bf Triaffine (Ours)} & 87.13 & 87.68 & 87.40 & 86.70 & 86.94 & 86.82 & 80.42 & 82.06 & \textbf{81.23} \\
{\bf Triaffine (Ours)} + ALBERT & 88.88 & 88.24 & \textbf{88.56} & 87.39 & 90.31 & \textbf{88.83} & - & - & - \\
\bottomrule
\end{tabular}
\caption{Results on the ACE2004, ACE2005, and GENIA datasets. BERT is the default encoder if not specified.}
\label{main results}
\end{table*}

\begin{table}[t]
\centering
\small
\begin{tabular}{lccc}
\toprule
Model + Encoder & \multicolumn{3}{c}{KBP2017}  \\
& P & R & $F_1$ \\
\midrule
ARN + LSTM & 77.7 & 71.8 & 74.6 \\
BiFlag + LSTM & 77.1 & 74.3 & 75.6 \\
Sequence to Set & 84.91 & 83.04 & 83.96  \\
Locate and Label & 85.46 & 82.67 & 84.05 \\
\midrule
{\bf Triaffine (Ours)} & 86.50 & 83.65 & 85.05 \\
{\bf Triaffine (Ours)} + ALBERT & \textbf{89.42} & \textbf{85.22} & \textbf{87.27}  \\
\bottomrule
\end{tabular}
\caption{Results on the KBP2017 dataset. BERT is the default encoder if not specified.}
\label{main kbp}
\end{table}

\subsection{Datasets}
We conduct our experiments on the ACE2004\footnote{\url{https://catalog.ldc.upenn.edu/LDC2005T09}}, ACE2005\footnote{\url{https://catalog.ldc.upenn.edu/LDC2006T06}} \cite{ace04}, GENIA \cite{kim2003genia} and KBP2017\footnote{\url{https://catalog.ldc.upenn.edu/LDC2019T12}} \cite{ji2017overview} datasets.
To fairly compare with previous works, we follow the same dataset split with \citet{lu2015joint} for ACE2004 and ACE2005 and use the split from \citet{lin2019sequence} for GENIA and KBP2017 datasets.
The statistics of all datasets are listed in Table \ref{dataset}.
Following previous work, we measure the results using span-level precision, recall, and $F_1$ scores.

\subsection{Implementation Details} \label{detail}
We use \texttt{BERT-large-cased} \cite{devlin-etal-2019-bert} and \texttt{albert-xxlarge-v2} \cite{lan2019albert} as the contextual embedding, \texttt{fastText} \cite{fasttext} as the word embedding in ACE2004, ACE2005 and KBP2017 dataset.
We use \texttt{BioBERT-v1.1} \cite{lee2020biobert} and \texttt{BioWordVec} \cite{zhang2019biowordvec}
as the contextual and word embedding in the GENIA dataset respectively.
We truncate the input texts with context at length 192.
The part-of-speech embeddings are initialized with dimension 50. The char embeddings are generated by a one-layer BiLSTM with hidden size 50.
The two-layers BiLSTM with a hidden size of 1,024 is used for the token representations.
For triaffine transformations, we use $d=256$ for the ACE2004, ACE2005, and KBP2017 dataset, and $d=320$ for the GENIA dataset, respectively.
We set $\mu_{aux}$ to $1.0$, and  select $m=30$ in both training and inference.
We use AdamW \cite{loshchilov2017decoupled} to optimize our models with a linear learning rate decay.
Detailed training parameters are presented in Appendix A.

\begin{table*}[ht]
\small
\centering
\begin{tabular}{lccccccccc}
\toprule
\multicolumn{8}{c}{Setting} & \multicolumn{2}{c}{Datasets} \\

&\multicolumn{3}{c}{Span Representation} & \multicolumn{4}{c}{Span Classification} & ACE2004 & GENIA \\
Setting &Label & Boundary & Function & Boundary & Attention & Cross & Function & \multicolumn{2}{c}{$F_1$}\\
\hline
(a)&$\times$ & $\times$ & $\times$ &  $\surd$ & $\times$ & $\times$ & bi. & 86.71  & 78.97 \\
(b)&$\surd$ & $\surd$ & tri. & $\times$ & $\surd$ & $\times$ & lin. & 87.36 & 80.50   \\
(c)&$\times$ & $\surd$ & tri. & $\surd$ & $\surd$ & $\times$ & tri. & 87.17  & 80.49   \\
(d)&$\surd$ & $\times$ & lin. & $\surd$ & $\surd$ & $\times$ & tri. & 87.14 & 80.50  \\
(e)&$\surd$ & $\surd$  & lin. & $\surd$ & $\surd$ & $\times$ & tri. & 87.35  & 80.63 \\
(f)&$\surd$ & $\surd$ & tri. & $\surd$ & $\surd$ & $\times$ & lin. & 87.49  & 80.70 \\
(g)&$\surd$ & $\surd$ & tri. & $\surd$ & $\surd$ & $\times$ & tri. & 87.54& 80.84\\
(h)&$\surd$ & $\surd$ & tri. & $\surd$ & $\surd$ & $\surd$ & tri. & \textbf{87.82} & \textbf{81.23} \\
\bottomrule
\end{tabular}
\caption{Ablation tests on ACE2004 development set and GENIA test set. Cross means using cross attention for span classification.
Lin. means linear transformation, bi. means biaffine transformation, and tri. means triaffine transformation.}
\label{main ablation}
\end{table*}

\subsection{Baselines}

\noindent\textbf{DYGIE} \cite{dygie} uses multi-task learning to extract entities, relations, and coreferences.

\noindent\textbf{MGNER} \cite{mgner} uses a detector to find span candidates and a classifier for categorization.

\noindent\textbf{BENSC} \cite{bensc} trains the boundary detection and span classification tasks jointly.

\noindent\textbf{TreeCRF} \cite{fu2021nested} views entities as nodes in a constituency tree and decodes them with a Masked Inside algorithm.

\noindent\textbf{Biaffine} \cite{yu2020named} classifies spans by a biaffine function between boundary representations.

\noindent\textbf{Pyramid} \cite{wang2020pyramid} designs pyramid layer and inverse pyramid layer to decode nested entities. 

We also report the results of models with other paradigms, including hypergraph-based methods \cite{sh}, transition-based methods \cite{mergelabel}, generative methods \cite{bartner,tan2021sequence,shen2021locate}, and so on.
We do not compare to BERT-MRC \cite{li2019unified} since they use additional resources as queries.
DYGIE++ \cite{wadden2019entity} and PURE \cite{zhong2021frustratingly} use different splits of the ACE datasets which are not comparable.

\subsection {Results}
We compare our method with baseline methods in Table~\ref{main results} for the ACE2004, ACE2005, and GENIA datasets and Table~\ref{main kbp} for the KBP2017 dataset, respectively.
With BERT as the encoder, our model achieves 87.40, 86.82, 81.23, and 85.05 scores in terms of $F_1$, outperforming all other span-based methods such as BENSC, Pyramid, TreeCRF, and Biaffine (+0.70 on ACE2004, +1.42 on ACE2005, +0.73 on GENIA). 
Compared with methods in other paradigms, our model also achieves the state-of-the-art results on the GENIA (+0.69 vs. Locate and Label) and KBP2017 dataset (+1.00 vs. Locate and Label) and shows comparable performances on ACE2004 (-0.01 vs. Locate and Label) and ACE2005 (-0.23 vs. Sequence to Set).
With a stronger encoder ALBERT, our model achieves 88.56, 88.83, and 87.27 scores in terms of $F_1$ on ACE2004, ACE2005, and KBP2017 respectively, which exceeds all existing baselines including the Pyramid model with ALBERT (+0.82 on ACE2004, +2.49 on ACE2005) and the previous state-of-the-art method on KBP2017 dataset (+3.22 vs. Locate and Label).

\subsection{Ablation Study}

Considering we leverage multiple factors in multiple parts of the model, we design the following ablation settings to validate the effectiveness of each factor and the proposed triaffine mechanism.

   \noindent(a) To show the effectiveness of triaffine mechanism, we use
   a baseline biaffine model with the combination of boundary representations:
    \begin{equation}
    p_{i,j,r} = 
    	\left[
    	\begin{aligned}
    	    \mathbf{h}_i \\
    	    1
    	\end{aligned}
    	\right]^T{\mathbf{V}_r}
    	\left[
    	\begin{aligned}
    	    \mathbf{h}_j \\
    	    1
    	\end{aligned}
    	\right]
    \end{equation}
    \noindent(b) To show the effectiveness of boundaries in scoring, we remove boundaries factor from scoring:
    \begin{equation}
        p_{i,j,r} = \mathbf{V}_r\mathbf{h}_{i,j,r} + \mathbf{b}_r
        \label{eq:linear score}
    \end{equation}
    \noindent(c) To show the effectiveness of labels in representation, we remove label factor in attention:
    \begin{equation}
        s_{i,j,k,r} = {\rm TriAff}(\mathbf{h}_i, \mathbf{h}_j, \mathbf{h}_k, {\mathcal W})
    \end{equation}
    \noindent(d) To show the effectiveness of boundaries in representation, we remove boundaries factor in attention:
    \begin{equation}
        s_{i,j,k,r} = s_{k,r} =  \mathbf{q}_r\cdot\mathbf{h}_k
    \end{equation}
    \noindent(e) To show the effectiveness of the triaffine mechanism in representations, we replace
    triaffine attention with linear attention:
    \begin{equation}
        s_{i,j,k,r} = \mathbf{W}_r(\mathbf{h}_i \mathbin\Vert \mathbf{h}_j \mathbin\Vert \mathbf{h}_k) + \mathbf{c}_r
    \end{equation}
    \noindent(f) To show the effectiveness of triaffine scoring, we replace
    triaffine scoring to linear scoring:
    \begin{equation}
        p_{i,j,r} = \mathbf{V}_r(
        \mathbf{h}_i \mathbin\Vert \mathbf{h}_j
        \mathbin\Vert
        \mathbf{h}_{i,j,r}) + \mathbf{b}_r
        \label{eq:linear score adv}
    \end{equation}
    \noindent(g) To show the effectiveness of cross-span interactions, we use 
    our partial model with intermediate predictions (model (a)-(g) use $p_{i,j,r}$). \\
    \noindent(h) Our full model (i.e, use $p^c_{i_l,j_l,r}$ as predictions).
    
\begin{table*}[ht]
\small
\centering
\begin{tabular}{p{6cm}ccccccccc}
\toprule
&& $p_{i,j,r}$ & & & $p_{i,j,r}^c$ \\
Span & Type & Probability& Rank & Type & Probability &&&&\\
\midrule
\multicolumn{10}{c}{... \textcolor{blue}{[}Cisco\textcolor{blue}{]\textsubscript{ORG}}'s been slammed, but once \textcolor{blue}{[}they\textcolor{blue}{]\textsubscript{ORG}}'re exposed to \textcolor{red}{[}the rest of \textcolor{blue}{[}the trading population\textcolor{blue}{]\textsubscript{PER}}\textcolor{red}{]\textsubscript{PER}} ...}\\
\midrule
Cisco & ORG &1.00 &1& ORG & 1.00 \\
they & ORG &1.00&2& ORG & 1.00 \\
the rest of the trading population & PER &1.00 &3& PER & 1.00 \\
the trading population & GPE & 0.50 &4& PER & 0.68 \\
population & None  & 1.00 &5& None  & 1.00 \\
\midrule
\multicolumn{10}{c}{... simian virus 40 enhancer activity was blocked by the \textcolor{blue}{[}MnlI-AluI fragment\textcolor{blue}{]\textsubscript{DNA}} in \textcolor{blue}{[}HeLa cells\textcolor{blue}{]\textsubscript{cl}} but not in \textcolor{blue}{[}B cells\textcolor{blue}{]\textsubscript{ct}}.}\\
\midrule
HeLa cells & cell line & 0.99 &1& cell line & 0.99 \\
B cells & cell type & 0.97 &2& cell type & 0.88 \\
MnlI-AluI fragment & DNA & 0.96 &3& DNA & 0.95 \\
simian virus 40 enhancer & DNA & 0.90 &4& DNA & 0.89  \\
MnlI-AluI & protein & 0.43 &5& None & 0.41 \\
40 enhancer & None & 0.99 &6& None & 1.00 \\
\bottomrule
\end{tabular}
\centering
\caption{Case study on ACE2004 and GENIA dataset.
Colored brackets indicate the boundaries and semantic types of entities in true labels. ``cl'' and ``ct'' is the abbreviation of \textit{cell line} and \textit{cell type}, respectively. }
\label{table:ex2}
\end{table*}

Table~\ref{main ablation} shows the results of ablation studies on ACE2004 and GENIA datasets. We use \texttt{BERT-large-cased} as the backbone encoder on ACE2004 and \texttt{BioBERT-v1.1} on GENIA, respectively.
By comparing (a) with (g), we observe significant performances drop (-0.87 on ACE2004, -1.87 on GENIA), which indicates that our proposed triaffine mechanism with multiple heterogeneous factors performs better than the biaffine baseline.
Comparing (b) with (g), we find that the boundary information contributes to span classification.
Comparing (c) and (d) with (g) supports that either label or boundary in the triaffine attention improves the performance.
The setting (g) performs better than (e) and (f), which shows the superiority of the triaffine transformation over the linear function.
We observe that (h) performs better than (g) (+0.28 on ACE2004, +0.39 on GENIA), proving the strength of triaffine attention with interactions among related spans.
The above studies support that our proposed triaffine mechanism with associated heterogeneous factors is effective for span representation and classification.

\subsection{Discussion}

\begin{figure}[t]
\centering
\includegraphics[width=2.9in]{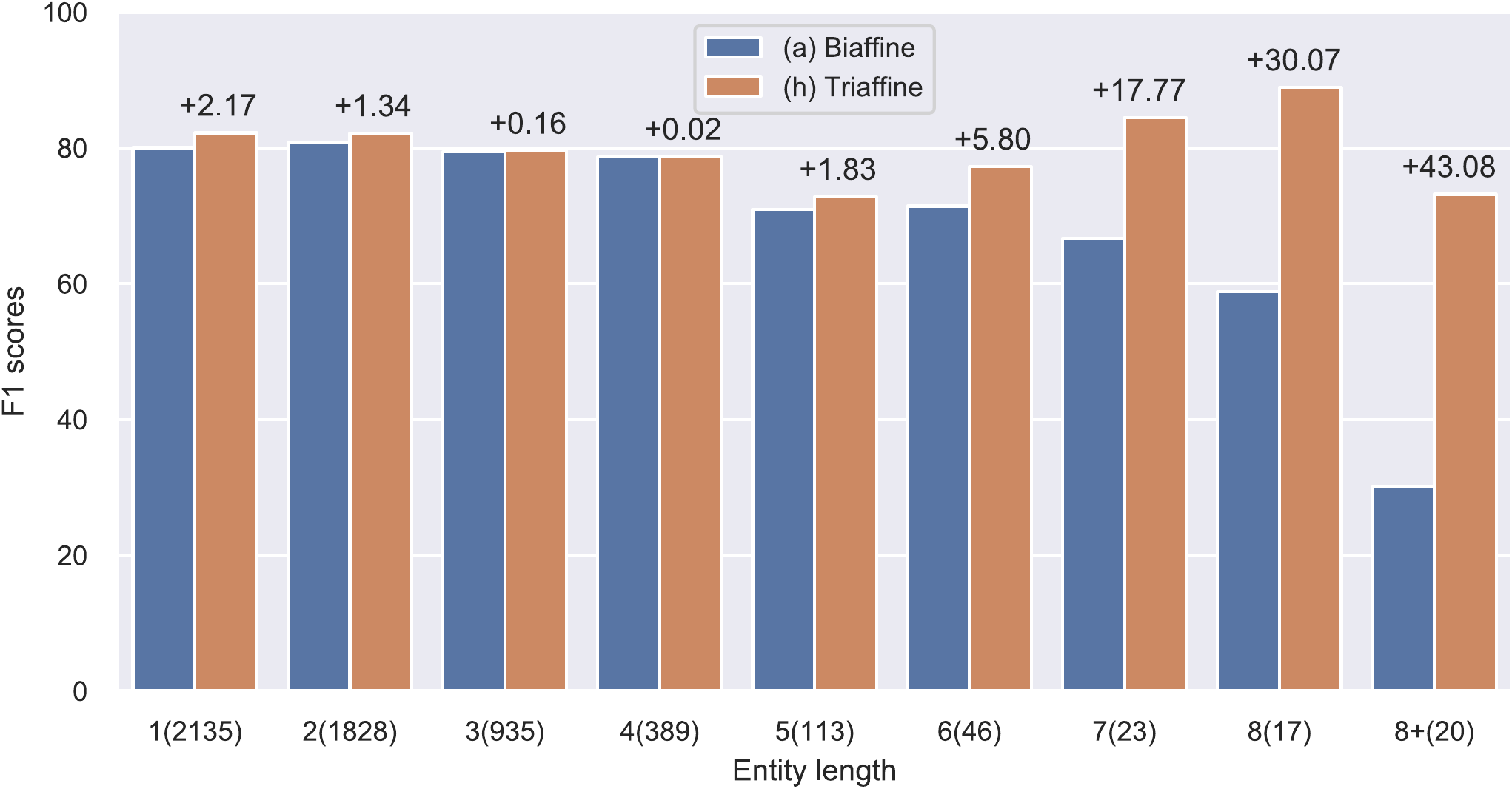}
\caption{Comparison between triaffine and biaffine models on GENIA with different lengths of entities. Entity counts are in the parentheses.}
\label{ablation:length}
\end{figure}

We compare the $F_1$ scores of GENIA between triaffine model (g) and biaffine model (a) grouped by entity lengths in Figure~\ref{ablation:length}. In all columns, the $F_1$ score of our method is
better than the baseline. Furthermore, the right columns show that the $F_1$ score of the baseline gradually decreases with the incremental entity lengths. However, our method based on the triaffine mechanism with heterogeneous factors takes advantage of the interaction from boundaries and related spans, which keeps consistent results and outperforms the baseline.

\begin{table}[h]
\small
\centering
\begin{tabular}{ccccc}
\toprule
 & \multicolumn{2}{c}{ACE2004} & \multicolumn{2}{c}{GENIA} \\
 & Flat  & Nested & Flat & Nested\\
  & (1,422)  & (1,092) & (4,307) &  (1,199)\\
\midrule
(a) & 88.51 & 84.19 & 80.09 & 74.23\\
(h) & 89.54 & 85.45 & 82.18 & 77.24\\
$\Delta$ & +1.03 & +1.26 & +2.09 & + 3.01 \\
\bottomrule
\end{tabular}
\centering
\caption{Comparison between triaffine and biaffine models on ACE2004 and GENIA grouped by flat or nested entities. Entity counts are in the parentheses.}
\label{table:nest}
\end{table}

The results grouped by flat or nested entities are shown in Table~\ref{table:nest}. Our method has consistent improvements than the baseline, especially for the nested setting. Based on the above observations, our method is good at solving long entities that are more likely to be nested, which supports our model is built upon the characteristics of nested NER.

\begin{figure}[t]
\centering
\includegraphics[width=3.0in]{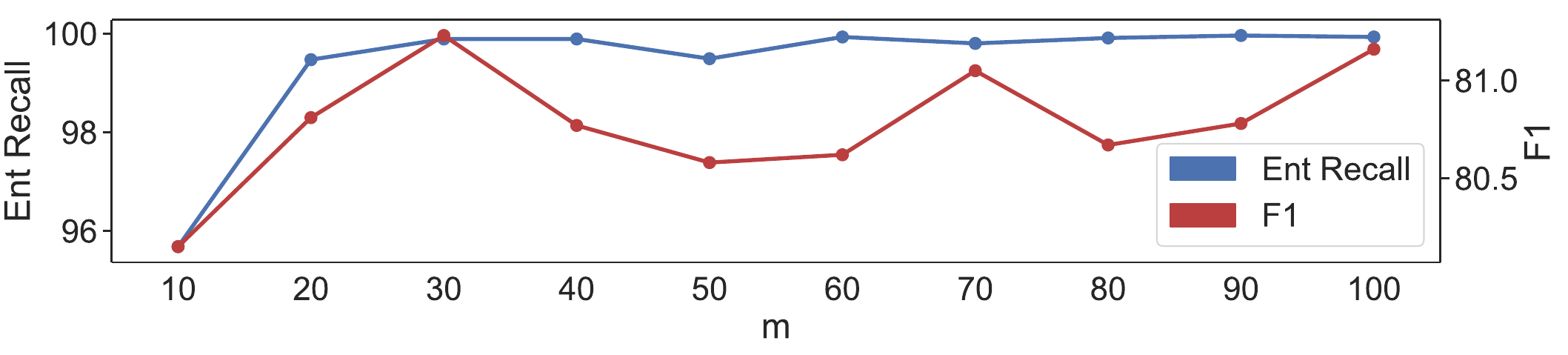}
\caption{Recall for entity spans and $F_1$ scores with different numbers of candidate spans in GENIA dataset.}
\label{ablation:genia span}
\end{figure}

At the stage of cross-span interactions, we only select top-$m$ spans in practice. In Figure \ref{ablation:genia span}, we analyze the number $m$ in two aspects. Firstly, we check the recall of entity spans.
We observe that taking top-30 spans achieves a recall of 99.89, which means it covers almost all entities. As the maximum number of entities is 25, we believe it is enough to select top-30 spans. Secondly, we check the model performance. With top-30 spans, the model achieves 81.23 scores in terms of $F_1$ and there is no obvious performance improvement with more candidates. Based on two above observations, we choose $m=30$, which can well balance the performance and efficiency.



Finally, we test the efficiency of the decomposition. 
Compared with the naive triaffine scoring that takes 638.1ms (509.4ms in GPU + 128.7ms in CPU), the decomposed triaffine scoring takes 432.7ms (330.5ms in GPU + 102.2ms in CPU) for 10 iterations, which leads to approximately 32\% speedup (details are shown in Appendix B).

\subsection{Case Study}

To analyze the effect of fusing information from related spans with the cross-span interaction, we show two examples from ACE2004 and GENIA datasets in Table~\ref{table:ex2}.
In the first example, the model first predicts ``\textit{the trading population}'' as ``GPE'', however, it revises to ``PER'' correctly by considering span interactions with the outer span ``\textit{the rest of the trading population}''.
In the second example, it first predicts ``\textit{MnlI-AluI}'' as ``protein''. By interacting with surrounding entities ``\textit{MnlI-AluI fragment}'', the model corrects its label to None. 

\section{Conclusion}


In this paper, we propose a span-based method for nested NER.
Heterogeneous factors including tokens, boundaries, labels, and related spans are introduced to improve span classification with a novel triaffine mechanism.
Experiments show our method outperforms all span-based methods and achieves state-of-the-art performance on four nested NER datasets.
Ablation studies show the introduced heterogeneous factors and triaffine mechanism are helpful for nested setting.
Despite that large-scale pretrained language models have shown consistent improvement over many NLP tasks, we argue that the well-designed features and model structures are still useful for complex tasks like nested NER. 
Furthermore, although we only verify our triaffine mechanism in nested NER, we believe it can also be useful in tasks requiring high order interactions like parsing and semantic role labeling.

\section*{Acknowledgements}
We would like to thank the anonymous reviewers for their helpful comments and suggestions.
We thank Yao Fu, Yongliang Shen, Shengxuan Luo, Hongyi Yuan, Zhengyun Zhao, Xu Chen, and Jiayu Li for their help.
This work was supported by Alibaba Group through Alibaba Research Intern Program.


\bibliography{anthology,custom}
\bibliographystyle{acl_natbib}

\appendix

\section{Reproducibility Checklist}
\label{appendix:para}
We set seeds of \textit{torch}, \textit{torch.cuda}, \textit{numpy}, and \textit{random} in Python to ensure reproducibility.
We use a grid search to find the best hyperparameters depending on development set performances.
We search contextual embedding learning rate among $\{$1e-5,3e-5$\}$. If the contextual embedding learning rate is 1e-5, we use static embedding learning rate and task learning rate as 1e-4 and 1e-5. If the contextual embedding learning rate is 3e-5, we use static embedding learning rate and task learning rate as 5e-4 and 3e-5. We search batch size among $\{$8,48,72$\}$.
We search MLP dropout ratio among $\{$0.1,0.2$\}$.
The final hyperparameters we used for four datasets are listed in Table~\ref{hyper para} and Table~\ref{hyper para albert}.

\begin{table}[ht]
\centering
\small
\begin{tabular}{lcccc}
\toprule
Parameters & ACE04 & ACE05 & KBP17 & GENIA \\
\midrule
Epoch & 50 & 50 & 50 & 15 \\
PLM lr & 1e-5 & 3e-5 & 1e-5 & 3e-5 \\
Static emb. lr & 1e-4 & 5e-4 & 1e-4 & 5e-4 \\
Task lr & 1e-5 & 3e-5 & 1e-5 & 3e-5 \\
$\sigma$ & 0.01 & 0.01 & 0.01 & 0.01 \\
Batch size & 8 & 72 &8 & 48 \\
$d$ & 256 & 256 & 256 & 320 \\
$m$ & 30 & 30 & 30 & 30  \\
Adam $\epsilon$ & 1e-8 & 1e-8 & 1e-8 & 1e-8\\
Warmup ratio & 0.0 & 0.0 & 0.0 & 0.0   \\
Emb. dropout & 0.2  & 0.2  & 0.2  & 0.2 \\
MLP dropout & 0.1 & 0.1 & 0.1 & 0.2 \\
Weight decay & 0.01 & 0.01 & 0.01 & 0.01 \\
Clipping grad & 0.1 & 0.1 & 0.1 & 0.1 \\
\bottomrule
\end{tabular}
\caption{Hyper-parameters for using BERT encoder.
}
\label{hyper para}
\end{table}

\begin{table}[ht]
\centering
\small
\begin{tabular}{lccc}
\toprule
Parameters & ACE04 & ACE05 & KBP17 \\
\midrule
Epoch & 10 & 10 & 10 \\
PLM lr & 1e-5 & 1e-5 & 3e-5 \\
Static emb. lr & 1e-4 & 1e-4  & 5e-4 \\
Task lr & 1e-5 & 1e-5  & 3e-5 \\
$\sigma$ & 0.01 & 0.01 & 0.01\\
Batch size & 8 & 8 & 72 \\
$d$ & 256 & 256 & 256 \\
$m$ & 30 & 30 & 30 \\
Adam $\epsilon$ & 1e-8 & 1e-8 & 1e-8  \\
Warmup ratio &  0.0 & 0.0 & 0.0 \\
Emb. dropout &0.2  & 0.2  & 0.2 \\
MLP dropout & 0.1 & 0.1  & 0.2 \\
Weight decay & 0.01 & 0.01 & 0.01 \\
Clipping grad &0.1 & 0.1 & 0.1 \\
\bottomrule
\end{tabular}
\caption{Hyper-parameters for using ALBERT encoder.
}
\label{hyper para albert}
\end{table}



\section{The Decomposition of Triaffine Scoring}
\label{appendix:decompose}
We introduce the decomposition of triaffine scoring in calculating $p_{i,j,r}$ and $p_{i,j,r}^c$.

The naive calculation procedure of $p_{i,j,r}$ is:
\begin{align}
    s_{i,j,k,r} &= {\rm TriAff}(\mathbf{h}_i, \mathbf{h}_j, \mathbf{h}_k, {\mathcal W}_r)
    \label{s} \\
    \alpha_{i,j,k,r} &= \frac{\exp(s_{i,j,k,r})}{\sum_{k'=i}^j \exp (s_{i,j,k',r})}
    \label{alpha} \\
    \mathbf{h}_{i,j,r} &= \sum_{k=i}^j\alpha_{i,j,k,r}{\rm MLP}(\mathbf{h}_k)
\label{attn} \\
    p_{i,j,r} &= {\rm TriAff}(\mathbf{h}_i, \mathbf{h}_j, \mathbf{h}_{i,j,r}, {\mathcal V}_r)
    \label{eq:slow}
\end{align}

For our proposed decomposition of $p_{i,j,r}$, we first calculate $\alpha_{i,j,k,r}$ as equations~\ref{s} and~\ref{alpha}. And we calculate:
\begin{align}
    o_{i,j,k,r} &= {\rm TriAff}(\mathbf{h}_i, \mathbf{h}_j, \mathbf{h}_k, {\mathcal V}_r)
    \label{eq:o} \\
    p_{i,j,r} &= \sum_{k=i}^j \alpha_{i,j,k,r}o_{i,j,k,r}
    \label{eq:decompose}
\end{align}
The main difference between naive calculation and decomposition calculation is between Equation~\ref{eq:slow} and Equation~\ref{eq:o}.

We suppose our batch size as $B$, sequence count as $N$, output dimensions of MLP layers as $d$, the count of spans for calculating cross span representations as $m$, and label count as $R$ (including None class). 
The shapes of tensors $[\mathbf{h}_i]$, $[\mathbf{h}_j]$, $[\mathbf{h}_k]$ are $B \times N \times d$.
The shape of tensor $[\mathbf{h}_{i,j,r}]$ is $B \times N \times N \times R \times d$.

\begin{table*}[t]
\centering
\small
\begin{tabular}{llcccc}
\toprule
Method & Function &\multicolumn{2}{c}{CPU Time}& \multicolumn{2}{c}{GPU Time} \\
&&Usage&Percentage&Usage&Percentage \\
\midrule
Equation~\ref{eq:slow} & \textit{aten::copy\_} & 0.5ms & 5.9\% & 223.7ms & 74.5\%  \\
      & \textit{aten::bmm} & 0.5ms & 5.0\% & 38.2ms & 12.7\% \\ 
      & \textit{aten::mm} & 1.5ms & 15.7\% & 37.1ms & 12.3\% \\
      & Total & 9.2ms & 100.0\% & 300.5ms & 100.0\% \\
\midrule
Equation~\ref{eq:o} & \textit{aten::copy\_} & 0.2ms & 4.7\% & 62.5ms & 42.9\% \\
      & \textit{aten::bmm} & 0.4ms & 10.0\% & 47.4ms & 32.6\% \\
      & \textit{aten::mm} & 0.3ms & 6.0\% & 34.4ms & 23.7\%  \\
      & Total & 4.4ms & 100.0\% & 145.6ms & 100.0\% \\
\midrule
Naive & \textit{aten::copy\_} & 7.3ms & 5.7\% & 302.3ms & 59.3\% \\
      & \textit{aten::bmm} & 1.2ms & 0.9\% & 109.3ms & 21.5\% \\ 
      & \textit{aten::mm} & 1.7ms & 1.4\% & 74.4ms & 14.6\% \\
      & \textit{aten::einsum} & 61.8ms & 48.0\% & 1.1ms & 0.2\% \\
      & \textit{aten::permute} & 36.7ms & 28.5\% & 0.8ms & 0.2\% \\
      & \textit{aten::reshape} & 1.3ms & 3.1\% & 0.5ms & 0.1\% \\
      & Total & 128.7ms & 100.0\% & 509.4ms & 100.0\% \\
\midrule
Decompose & \textit{aten::copy\_} & 0.7ms & 0.8\% & 136.7ms & 41.4\% \\
      & \textit{aten::bmm} & 1.2ms & 1.2\% & 102.6ms & 31.0\% \\ 
      & \textit{aten::mm} & 5.4ms & 5.3\% & 69.0ms & 20.9\% \\
      & \textit{aten::einsum} & 32.0ms & 31.3\% & 1.1ms & 0.3\% \\
      & \textit{aten::permute} & 15.4ms & 15.1\% & 0.7ms & 0.2\% \\
      & \textit{aten::reshape} & 37.4ms & 36.6\% & 0.5ms & 0.2\% \\
      & Total & 102.2ms & 100.0\% & 330.5ms & 100.0\% \\
\bottomrule      
\end{tabular}
\caption{Time usage compared with naive triaffine scoring and decomposed triaffine scoring.}
\label{table:compare1}
\end{table*}

We benchmark the performances of Equation~\ref{eq:slow} and Equation~\ref{eq:o} in PyTorch for 10 iterations.
We use the same hyper-parameters and devices as our main experiments.
We levearge opt\_einsum\footnote{\url{https://github.com/dgasmith/opt_einsum}} to calculate triaffine transformations in both equations.

Table~\ref{table:compare1} shows the time usage comparison between Equation~\ref{eq:slow} and Equation~\ref{eq:o}.
Equation~\ref{eq:slow} uses 309.7ms (300.5ms in GPU + 9.2ms in CPU) and Equation~\ref{eq:o} uses 150.1ms (145.6ms in GPU + 4.4ms in CPU).
The larger tensor size and higher rank of $[\mathbf{h}_{i,j,r}]$ results in slower calculations of \textit{aten::bmm}, \textit{aten::copy\_} and \textit{aten::permute} in Equation~\ref{eq:slow}.
The time usage differences are clearly dominated by the function \textit{aten::copy\_}, which is optimized by our decomposition.

We also compare the time usage between the naive triaffine scoring and the decomposed triaffine scoring in Table~\ref{table:compare1}.
The naive triaffine scoring takes 638.1ms (509.4ms in GPU + 128.7ms in CPU), and the decomposed triaffine scoring takes 432.7ms (330.5ms in GPU + 102.2ms in CPU) for 10 iterations, which leads to approximately 32\% speedup.
The GPU time usages are reasonable since they both need to calculate two triaffine transformations.
The CPU time usages increase for both naive and decomposition triaffine scoring.
Additional CPU time usages come from function \textit{aten::einsum}, \textit{aten::permute}, and \textit{aten::reshape}, and the naive calculation increases more due to slower \textit{aten::einsum}.
Overall, the decomposition triaffine scoring uses less time on both GPU and CPU than the naive triaffine scoring.

Futhermore, we also test the time usage of $p_{i,j,r}^c$ using two calculation procedures. We find using the decomposition triaffine scoring still has about 6\% speed up (naive:125.8ms in GPU + 15.0ms in CPU vs. decomposition:115.5ms in GPU + 16.8ms in CPU) regardless the relatively small size of $\mathbf{h}_{i,j,r}^c$ (The shape of tensor $[\mathbf{h}_{i,j,r}^c]$ is $B \times m \times R \times d$).

\end{document}